\documentclass[letterpaper, 10 pt, journal, twoside]{ieeetran}  

\IEEEoverridecommandlockouts                              

\usepackage{graphics} 
\usepackage{epsfig} 
\usepackage{times} 
\usepackage{amsmath} 
\usepackage{amssymb}  
\usepackage{amsfonts}
\usepackage[colorlinks,linkcolor=black,anchorcolor=black,citecolor=black,urlcolor=black]{hyperref}
\usepackage{booktabs}
\usepackage{multirow}
\usepackage{graphicx}
\usepackage{subfigure}
\usepackage{cite}

\graphicspath{{figures/}}

\title{AutoTrans: A Complete Planning and Control Framework for Autonomous UAV \\Payload Transportation}

\author{Haojia Li$^{2}$, Haokun Wang$^{2}$, Chen Feng$^{2}$, Fei Gao$^{*3,4}$, Boyu Zhou$^{*1}$, and Shaojie Shen$^{2}$

\thanks{Manuscript received: April, 24, 2023; Revised: July, 13, 2023; Accepted: August, 9, 2023. This paper was recommended for publication by Editor Pauline Pounds upon evaluation of the Associate Editor and Reviewers’ comments.
This work was supported by The Research Grants Council General Research Fund (RGC GRF) project RMGS20EG20} 
\thanks{$^{1}$School of Artificial Intelligence, Sun Yat-Sen University, Zhuhai, China.}%
\thanks{$^{2}$Department of Electronic and Computer Engineering, Hong Kong University of Science and Technology, Hong Kong, China.}%
\thanks{$^{3}$State Key Laboratory of Industrial Control Technology, Institute of Cyber-Systems and Control, Zhejiang University, Hangzhou, China.}%
\thanks{$^{4}$Huzhou Institute, Zhejiang University, Huzhou, China.}%
\thanks{*Corresponding authors: Boyu Zhou, Fei Gao}
\thanks{Email: {\tt\footnotesize hlied@ust.hk, zhouby23@mail.sysu.edu.cn}}
\thanks{Digital Object Identifier (DOI): 10.1109/LRA.2023.3313010}
}

\begin{document}

\maketitle


\begin{abstract}

    The robotics community is increasingly interested in autonomous aerial transportation. Unmanned aerial vehicles with suspended payloads have advantages over other systems, including mechanical simplicity and agility, but pose great challenges in planning and control. To realize fully autonomous aerial transportation, this paper presents a systematic solution to address these difficulties. First, we present a real-time planning method that generates smooth trajectories considering the time-varying shape and non-linear dynamics of the system, ensuring whole-body safety and dynamic feasibility. Additionally, an adaptive NMPC with a hierarchical disturbance compensation strategy is designed to overcome unknown external perturbations and inaccurate model parameters. Extensive experiments show that our method is capable of generating high-quality trajectories online, even in highly constrained environments, and tracking aggressive flight trajectories accurately, even under significant uncertainty. We plan to release our code to benefit the community.

\end{abstract}


\section{Introduction}
\IEEEPARstart{R}{ecently}, robotics researchers have shown a growing interest in aerial transportation, driven by the need for efficient, safe, and flexible transportation in extensive scenarios, e.g., disaster relief, logistics, and agriculture. A typical aerial transportation system employs an active manipulator\cite{fishman2021Dynamica} or a cable\cite{sreenath2013TrajectoryGenerationControlc} attached to the aerial vehicle. Compared to active manipulators, unmanned aerial vehicles with suspended payload systems (UAV payload systems) are preferred as it needs no additional actuation, which reduces mechanical complexity and weight, allowing a higher load budget \cite{foehn2017FastTrajectoryOptimization}. Furthermore, the UAV's attitude dynamics are less affected by the payload than a manipulator, allowing improved attitude control and potentially more agile maneuvers \cite{sreenath2013GeometricControlDifferentiala}.

Although UAV payload systems offer several benefits, they also pose significant challenges. One major issue is the lack of a motion planner that meets three essential criteria: real-time planning, dynamic feasibility, and whole-body safety. Particularly, the algorithm must efficiently account for the system's time-varying shape and enable safe yet agile maneuvers in complex unknown environments. Some existing methods \cite{son2019Realtime,foehn2017FastTrajectoryOptimization,potdar2020Online} avoid collisions by using the system's bounding sphere or simplified obstacle shapes, which are either conservative or impractical in complex scenes. Although some works consider the exact shape of the system\cite{zeng2020Differential}, they are not real-time capable due to the complex system model.

\begin{figure}[t]
    \centering
    \includegraphics[width=0.85\linewidth]{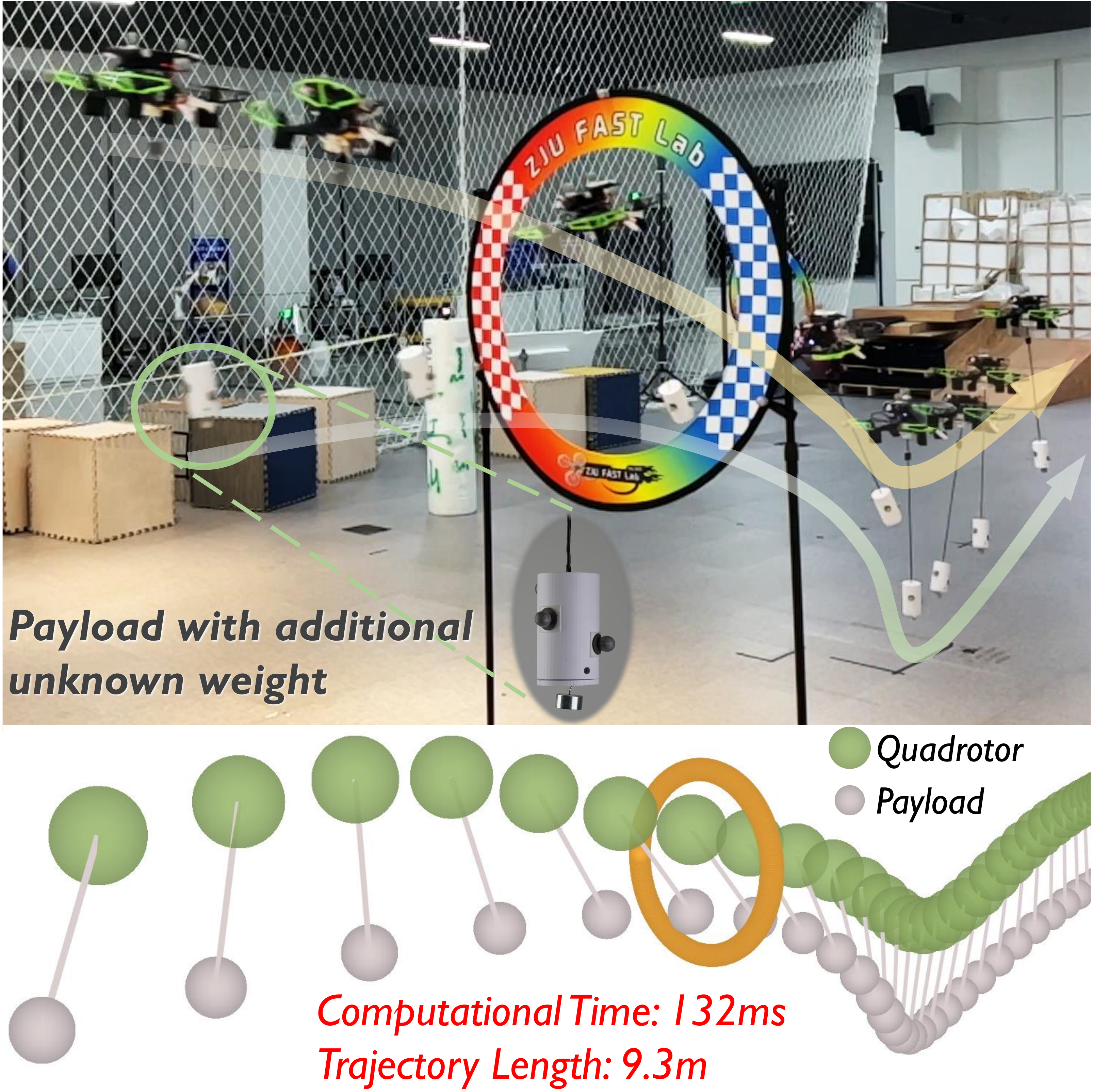}
    \caption{A quadrotor with a suspended payload system is passing a narrow circle. An unknown weight (50g) hangs below the payload. The trajectory generation time is 132ms with 9.3m length on quadrotor's onboard computer.}
    \label{fig:pass_gate}
    \vspace{-0.5cm}
\end{figure}

Another challenge is the lack of a robust and adaptive controller that accurately tracks trajectories under significant uncertainty. External disturbances such as wind or rotor drag, along with inaccuracies in modeling (e.g., assuming that the cable is attached to UAV's center of mass), can all affect control accuracy. Moreover, extra uncertainty would be introduced for more flexible transportation. Specifically, it is desirable that the system can transport various payloads adaptively without knowing the precise weight of a payload. This enables efficient load-and-fly transportation but also introduces uncertain parameters into the system model.

To address these challenges, this paper presents a systematic solution for fully autonomous UAV payload transportation. Firstly, we introduce a real-time planning solution to generate smooth trajectories. We exploit the differential flatness property of UAV payload system to derive the system's states, thereby converting the whole-body safety and dynamic feasibility constraints into continuous and differentiable penalty functions. The problem is then formulated as an unconstrained optimization problem that can be efficiently solved by a spatial-temporal trajectory optimizer. To deal with strong nonlinearity, we further adopt a kinodynamic search front-end to find promising initial guesses. Secondly, to overcome unknown external perturbations as well as inaccurate model parameters, we propose an adaptive NMPC with a hierarchical disturbance compensation strategy. An online external force estimator is designed, which compensates for disturbance forces acting on both UAV and payload. We further apply Incremental Nonlinear Dynamic Inversion (INDI) to the angular velocity controller, compensating unknown torque. The proposed method can generate high-quality trajectories online even in highly constrained environments, such as the one in \autoref{fig:pass_gate}. Meanwhile, our controller tracks trajectories robustly and accurately, even with additional unknown payloads.

In summary, our work has three contributions:

\begin{enumerate}
	\item { A real-time differential flatness-based spatio-temporal planning method, which generates smooth trajectories that ensure whole-body safety and dynamic feasibility.  }
	\item{ A robust NMPC controller with a hierarchical disturbance compensation strategy, resisting perturbations and inaccurate model parameters. }
	\item{ Extensive simulation and real-world experiments for validating our methods. We release our code, including planning, controller, and simulation\footnote[1]{\href{https://github.com/SYSU-STAR/AutoTrans}{https://github.com/SYSU-STAR/AutoTrans}}. }
\end{enumerate}

\section{Related Work}

Numerous planning and control methods have been proposed to address the unique challenges presented by UAVs with suspended payloads.
As for trajectory generation,
\cite{palunko2012Trajectory,palunko2012Agile,sreenath2013TrajectoryGenerationControlc} generated minimum swing trajectory, but did not consider obstacle avoidance. \cite{silveira2020Aggressive} used RRT, which can not ensure energy efficiency. In \cite{tang2015Mixed}, mixed integer quadratic programming was used to generate narrow-gap-passing trajectories and demonstrated load throwing. \cite{zeng2020Differential,foehn2017FastTrajectoryOptimization} achieved maneuvers with complementary constraint. While the referred work demonstrated promising results, it was hard to generate collision-free trajectories in real time. Alternatively, \cite{son2019Realtime,son2020Realtime,potdar2020Online} achieved real-time trajectory generation, but they all conservatively assumed that obstacles are ellipsoid, which may not generalize well to complex and unstructured environments.

As for controllers, some researchers
\cite{sreenath2013GeometricControlDifferentiala,goodarzi2014GeometricStabilizationQuadrotor} proposed geometric controllers, which employ a coordinate-free dynamic model derived from the properties of Lie groups with a cascade structure. However, this type of controller was unable to track both the UAV and the suspended payload simultaneously. Some approaches, such as \cite{guerrero2015PassivityBasedControl,guerrero-sanchez2017Swingattenuation,palunko2013Reinforcement}, focus on minimizing the swing angle of the suspended payload to achieve stabilization. However, these approaches tend to restrict the range of motion of the payload, thereby limiting its ability to perform high-speed agile maneuvers. All the above methods did not consider the disturbance or the uncertainty, which is essential for executing tasks outdoors, e.g., medical rescue. \cite{pereira2021Aerial} did consider wind disturbances with a complex model, but this approach had strong assumptions about the environment and did not demonstrate its performance in real-world scenarios.  

Despite the significant progress, there are still several challenges that need to be addressed to apply UAVs with suspended payloads in the real world. These include developing methods that can handle complex and cluttered environments, improving the robustness of control methods to disturbances and uncertainties, and reducing the computational complexity of planning and control methods to enable real-time operation.

\section{Preliminaries}
This section introduces the dynamics model and differential flatness for a quadrotor with a cable-suspended payload.
\subsection{Nominal System Dynamics}
\label{sec:nominalsystemdynamics}
Quadrotor with cable-suspended payload has two modes depending on the cable tension. One is the taut mode, and the other one is the slack mode. In transportation tasks, we desire the UAV payload system to be energy-efficient. However, the energy consumption of hybrid slack and taut mode is typically higher than always-taut mode due to the large acceleration required to compensate when the cable becomes taut again \cite{foehn2017FastTrajectoryOptimization}. Therefore, we assume the cable is always taut in this paper. The whole system can be considered as a rigid body, and the cable can be simplified as a massless rod. 

The system's configuration is illustrated as \autoref{fig:payload_system}. Similar to \cite{sreenath2013GeometricControlDifferentiala,li2021PCMPCPerceptionConstrainedModelc}, the system dynamics is formulated as
\begin{equation}
    ( m_{Q} +m_{L})\left(\ddot{\mathbf{x}}_{L} +g\mathbf{e}_{z}^{w}\right)  = \left(\boldsymbol{\rho } \cdot f\mathbf{Re}_{z}^{w} -m_{Q} l(\dot{\boldsymbol{\rho }} \cdotp \dot{\boldsymbol{\rho }})\right)\boldsymbol{\rho },
\end{equation}
\begin{equation}
    m_{Q} l\ddot{\boldsymbol{\rho }} +m_{Q} l(\dot{\boldsymbol{\rho }} \cdotp \dot{\boldsymbol{\rho }})\boldsymbol{\rho } =\boldsymbol{\rho } \times \left(\boldsymbol{\rho } \times f\mathbf{Re}_{z}^{w}\right)  ,
\end{equation}
\begin{equation}
    \label{equ:xlxqtrans}
    \mathbf{x}_{Q} = \mathbf{x}_{L}-l\boldsymbol{\rho },
\end{equation}
\begin{equation}
\dot{\mathbf{R}} =\mathbf{R}\hat{\boldsymbol{\omega }}, \ \boldsymbol{\tau}=\mathbf{J\dot{\boldsymbol{\omega }}} +\boldsymbol{\omega } \times \mathbf{J}\boldsymbol{\omega },
\end{equation}
where $\mathbf{x}_{L}$ and $\mathbf{x}_{Q}$ are center position of the payload and quadrotor. $\ddot{\mathbf{x}}_{L}$ is payload's acceleration. $\boldsymbol{\rho }$ is cable direction (a unit vector from quadrotor to payload center). $l$ is cable length. $\mathbf{J}$ is the quadrotor's inertia matrix. $\boldsymbol{\omega }$ is angular velocity of the quadrotor. $f$ is quadrotor thrust. $\mathbf{R}$ is rotation of quadrotor. $\boldsymbol{\tau}$ is moment vector of quadrotor. $\mathbf{e}_{z}^{w}$ is the z-axis of world frame. 

\subsection{Differential Flatness}
\label{sec:flatness}
  Differential flatness allows us to express the system's state and control inputs in terms of these flat outputs, reducing the optimization problem's dimension and enabling efficient trajectory planning and control\cite{wang2022GeometricallyConstrainedTrajectorya}. In the case of a quadrotor with a cable-suspended payload, this property holds when the cable is taut. The flat outputs $\mathcal{Z}=[\mathbf{x}_{L}, \psi ]$ for this system are payload's position $\mathbf{x}_{L}$ and the quadrotor's yaw angle $\psi$ \cite{sreenath2013TrajectoryGenerationControlc}. 
\begin{equation}
    \label{equ:flat_rho}
    m_{L}\left(\ddot{\mathbf{x}}_{L} +g\mathbf{e}_{z}^{w}\right) =-f_{c}\boldsymbol{\rho }, \  \boldsymbol{\rho } =-\frac{\left(\ddot{\mathbf{x}}_{L} +g\mathbf{e}_{z}^{w}\right)}{\left\Vert \left(\ddot{\mathbf{x}}_{L} +g\mathbf{e}_{z}^{w}\right)\right\Vert _{2}^{2}}
\end{equation}
\begin{equation}
    \label{equ:flat_rot}
    f\mathbf{Re}_{z}^{w} =m_{Q}\left(\ddot{\mathbf{x}}_{Q} +g\mathbf{e}_{z}^{w}\right) -f_{c}\boldsymbol{\rho }.
\end{equation}
\begin{figure*}
    \centering
    \includegraphics[width=0.85\linewidth]{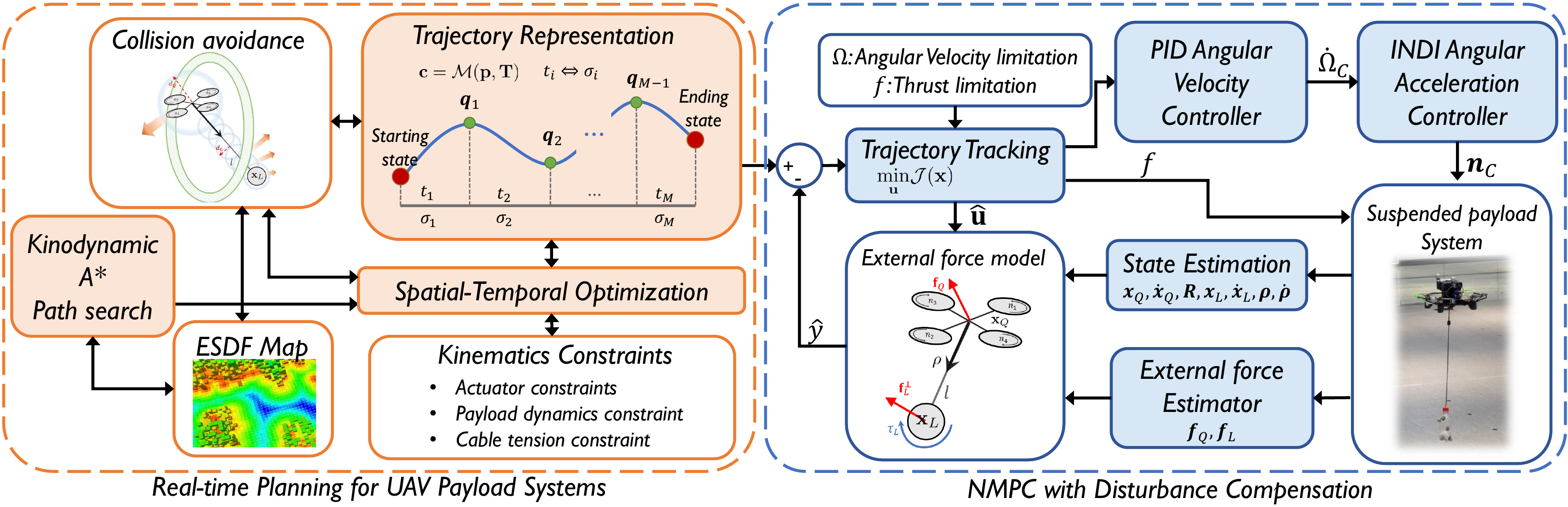}
    \caption{System Overview. Kinodynamic A* obtains the initial path, and spatio-temporal optimization generates a smooth, feasible, collision-free trajectory. NMPC with force estimator and INDI handles disturbances during tracking.}
    \label{fig:architecture}
    \vspace{-0.5cm}
\end{figure*}
The position $\mathbf{x}_{Q}$ and its derivatives for the quadrotor can be determined using equation \eqref{equ:xlxqtrans}. The z-axis of the quadrotor can be obtained from equation \eqref{equ:flat_rho} and equation \eqref{equ:flat_rot}. As $\mathbf{x}_{Q}$ and $\psi$ are known along with their derivatives, the remaining states and control inputs can be obtained using the conventional quadrotor method, as described in \cite{mellinger2011MinimumSnapTrajectory}.

\subsection{System Overview}
The overall architecture is depicted in \autoref{fig:architecture}. The kinodynamic A* obtains a collision-free path as promising initial guesses. Then, a spatio-temporal trajectory optimization method is utilized to generate a smooth, dynamically feasible, and collision-free trajectory. The generated polynomial trajectory is tracked by a nonlinear model predictive controller (NMPC) incorporating a hierarchical disturbance compensation strategy that includes INDI control for compensating angular accelerations and a force estimator for compensating external forces. 

\section{Real-time Planning for UAV Payload Systems}
In this section, we introduce the differential flatness-based spatio-temporal optimization formulation.

\subsection{Nonlinear Optimization Construction}
We utilize $M$ pieces of $D$-dimension polynomial with degree $N=2s-1$ to represent the whole trajectory. Each piece is expressed as $\mathcal{Z}_{i} (t)=\mathbf{c}_{i}^{T}\mathbf{\beta }( t) ,\ \mathbf{\beta }( t) =\left[ 1,t_i,...,t_i^{2s-1}\right]^{T}$.
In our case, we choose $D=3$ and $N=7$ to represent the payload position $\mathbf{x_L}$ in flat output space. 

The objective function minimizes the control energy involving a time regularization written as \eqref{equ:opt_cost}, where $Q_{J}$ and $\lambda_T$ are the weight of energy and time penalty, respectively. 
\begin{subequations}
    \begin{align}
        \underset{\mathbf{c,T}}{\min} J&(\mathbf{c,T} ) =\underbrace{\sum _{i=1}^{M}\int _{0}^{t_{i}}\mathcal{Z}_{i}^{( s)}( t)Q_{J}\mathcal{Z}_{i}^{( s)}( t)dt}_{\mathrm{Energy\ Cost}} +\underbrace{\lambda _{T} \| \mathbf{T} \| }_{\mathrm{Time\ Cost}}     \label{equ:opt_cost}\\
        s.t.\ &\mathcal{Z}_{1}^{[ s-1]}( 0) =\mathbf{z}_{0} ,\ \mathcal{Z}_{M}^{[ s-1]}( t_{M} ) =\mathbf{z}_{f} ,\label{equ:opt_boundary_conditions}\\
          &\mathcal{Z}_{i}^{[ s-1]}( t_{i}) =\mathcal{Z}_{i+1}^{[ s-1]}( 0), \label{equ:opt_continuous_constraint}\\
          &t_i > 0, t_i \in \mathbf{T} , \label{equ:time_positive}\\
          &\mathcal{C}_{b}\left(\mathcal{Z}_{i}( t) ,\mathcal{Z}_{i}^{( 1)}( t) ,...,\mathcal{Z}_{i}^{( s)}( t)\right) \leqslant 0, b \in \mathcal{G}.\label{equ:constraints}
    \end{align}
\end{subequations}
The dynamics constraints and collision avoidance are expressed as violation function $\mathcal{C}_{b}$ \eqref{equ:constraints} where $\mathcal{G}$ is the set of all constraints. The equation \eqref{equ:opt_boundary_conditions} and \eqref{equ:opt_continuous_constraint} denote boundary conditions and continuous constraints of trajectory separately where $\mathcal{Z}_{i}^{(s)}$ is $s$ order derivative on flat output $\mathcal{Z}_{i}$, and $\mathcal{Z}_{i}^{[ s-1]}$ is all derivative from $0$-order to $s-1$ order.

To deform the spatio-temporal trajectory, we utilize the MINCO representation \cite{wang2022GeometricallyConstrainedTrajectorya}. This approach allows us to optimize the time and shape of the trajectory simultaneously while naturally satisfying the boundary conditions \eqref{equ:opt_boundary_conditions} and continuous constraints \eqref{equ:opt_continuous_constraint}. Instead of direct parametrization by coefficients, $\mathbf{c}$, MINCO represents a polynomial spline with initial, terminal states, intermediate points $\mathbf{p}$ and time allocation $\mathbf{T}$. The spline guarantees to pass through $\mathbf{p}$ within a specific time $\mathbf{T}$. We transfer the optimization space by $\mathbf{c} = \mathcal{M}(\mathbf{p}, \mathbf{T})$ to reduce the optimization variables and improve numerical stability, where $\mathcal{M}$ expresses a smooth and linear-complexity map. Furthermore, $\mathcal{M}$ can be treated as a differentiable function\cite{wang2022GeometricallyConstrainedTrajectorya}. 

\subsection{Whole-body Collision Avoidance}
\label{sec:collision_avoidance}
\begin{figure}[t]
    \centering
    \includegraphics[width=0.8\linewidth]{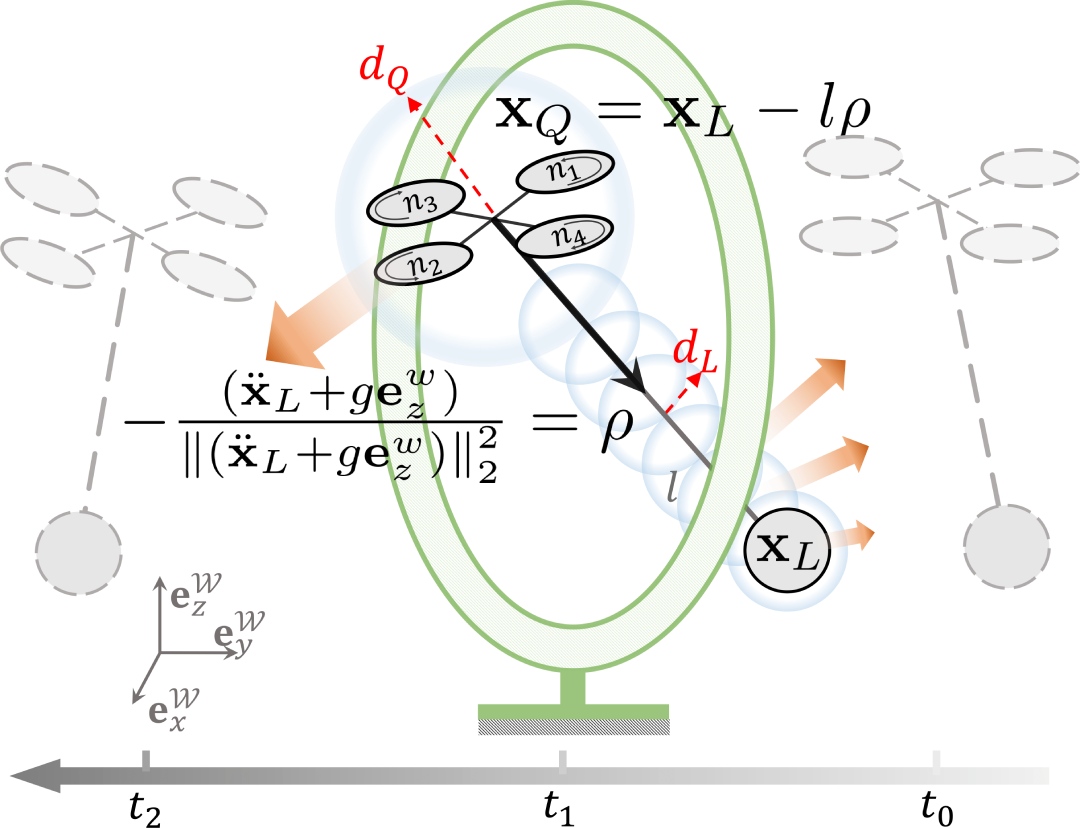}
    \caption{Configuration of the UAV payload system. The system shape is simplified as a bunch of bubbles. When bubbles collide with obstacles, the ESDF field pulls the bubbles away from the obstacles.}
    \label{fig:payload_system}
    \vspace{-0.5cm}
\end{figure}
The payload transportation system comprises multiple rigid bodies, leading to variations in its shape with changing state. Representing the entire system as a single point reduces the solution space, making it difficult for the system to pass through narrow gaps. To overcome this, we approximate the shape of the payload transportation system using a set of spherical bubbles (as shown in Figure \ref{fig:payload_system}). The system position and shape are decided by the quadrotor and payload position. Based on the differential flatness property, \eqref{equ:xlxqtrans} and \eqref{equ:flat_rho}, the system position and shape are determined by payload position $\mathbf{x}_L$ and acceleration $\ddot{\mathbf{x}}_L$. We ensure that the distance between these bubbles and obstacles is greater than the safe distance $d_s$. We formulate collision avoidance constraints, as \eqref{equ:collision_avoidance_violation},
    \begin{align}
    &d( l_{s}) =d_{s} -\mathbf{E}(\mathbf{x}_{L} -l_{s}\boldsymbol{\rho }), \label{equ:collision_avoidance_distance} \\
    &\mathcal{C}_{bc} =\frac{1}{N}\sum _{j=0}^{N}\max\left[ \left(d\left(\frac{j}{N} l\right)\right)^{3} ,0\right] \leqslant 0,\label{equ:collision_avoidance_violation}
\end{align}
where $N$ is the number of bubbles. The $d_{s}$ can be varying according to system shape with different $j$. The safe distance of payload and cable is $d_L$. The $d_Q$ represents the quadrotor's safe distance. We maintain a discretization euclidean signed distance field (ESDF) $\mathbf{E}$ for querying the shortest distance to obstacles with \cite{felzenszwalb2012distance}. When the quadrotor and payload are close to the obstacle, the gradient will push the shape out of the obstacle like the orange arrows in \autoref{fig:payload_system}. In practice, trilinear interpolation is utilized to improve the precision of distance and gradient. 

\subsection{Dynamic Constraints} 
To ensure the feasibility of the trajectory, we impose dynamic constraints on the system using differential flatness.

\subsubsection{Quadrotor actuator constraints}
We consider hardware limitations by limiting the maximum thrust and tilt angle of the quadrotor. From the flatness theorem in section \ref{sec:flatness}, the thrust and the quadrotor z-axis is written as \eqref{equ:thrustandrotation},
\begin{equation}
    \label{equ:thrustandrotation}
    f\mathbf{Re}_{z}^{w} =( m_{Q} +m_{L})\left(\ddot{\mathbf{x}}_{L} +g\mathbf{e}_{z}^{w}\right) -m_{Q} l\ddot{\boldsymbol{\rho }}.
\end{equation} 
We limits the thrust within the boundaries $f_{l}$ and $f_{u}$ by
\begin{equation}
    \label{equ:thrustconstraint}
    \mathcal{C}_{bf}(\mathbf{z}) =\left( f-\frac{f_{u}  + f_{l}}{2}\right)^{2} - \left(\frac{f_{u}  - f_{l}}{2}\right)^{2} \leqslant 0.
\end{equation} 
Limiting the thrust benefits quadrotor attitude stabilization and avoiding unfeasible trajectory.

The tilt angle of the quadrotor should not exceed the limitation to ensure dynamic feasibility and prevent the quadrotor from flipping. We limit the tilt angle of the quadrotor to $\theta_{max} = 60^\circ$. The tilt angle constraint is defined as equation \eqref{equ:tiltangleconstraint},
\begin{equation}
    \label{equ:tiltangleconstraint}
    \mathcal{C}_{b\theta}(\mathbf{z}) =\cos( \theta _{max}) -\mathbf{e}{_{z}^{w}}^{T}\mathbf{Re}_{z}^{w} \leqslant 0.
\end{equation} 

\subsubsection{Payload dynamics constraint}
We restrict the payload velocity and acceleration in a reasonable magnitude by \eqref{equ:payloaddynamicsconstraint},  
\begin{equation}
    \label{equ:payloaddynamicsconstraint}
    \mathcal{C}_{bp}(\mathbf{z}) =\begin{cases}
        \| \dot{\mathbf{x}}_{L} \| _{2}^{2} -v_{max}^{2} \leqslant 0\\
        \| \ddot{\mathbf{x}}_{L} \| _{2}^{2} -a_{max}^{2} \leqslant 0
        \end{cases}.
\end{equation} 

\subsubsection{Cable tension constraint}
As discussed in \ref{sec:nominalsystemdynamics}, We assume the cable is always taut. From equation \eqref{equ:flat_rho}, the cable tension is expressed as $f_{c}=\Vert m_{L}\left(\ddot{\mathbf{x}}_{L} +g\mathbf{e}_{z}^{w}\right)\Vert_{2}$. As we project the force on the z-axis, the cable tension constraint can be written as equation \eqref{equ:cabletensionconstraint}, 
\begin{equation}
    \label{equ:cabletensionconstraint}
    \mathcal{C}_{bf_c}(\mathbf{z}) = \epsilon -\ddot{\mathbf{x}}_{L}\mathbf{e}_{z}^{w} - g  \leqslant 0, 
\end{equation} 
where $\epsilon$ is a small positive number to avoid the numerical precision problem. 

\subsection{Unconstrained Optimization}
To speed up optimization, we hope to eliminate all the constraints and transfer the original problem into an unconstrained problem. Considering the time of each segment, it should be strictly positive as \eqref{equ:time_positive}. We introduce an unconstrained virtual time $\sigma  = [\sigma_1, ..., \sigma_i, ..., \sigma_n]$ and use a diffeomorphism map\cite{wang2022GeometricallyConstrainedTrajectorya} 
to map $\sigma_i \in \mathbb{R}$ to the real duration $t_i \in \mathbb{R}^+$. 
 
We adopt the integration to compute the constraint violation to reduce the infinite number of constraints to a finite number, 
\begin{equation}
    \label{equ:discreteintegration}
    S=\sum _{i=1}^{M}\int _{0}^{t_{i}} L_{1}(\mathcal{C}_{b}( \mathbf{z})) dt, 
\end{equation}
where $L_1(\cdot)$ is smooth penalty function\cite{han2022DifferentialFlatnessBasedTrajectory}.
In summary, the trajectory optimization problem can be redefined as an unconstrained optimization problem shown as equation \eqref{equ:unconopt}, 
\begin{equation}
    \label{equ:unconopt}
    \underset{\mathbf{q,\sigma }}{\min} J(\mathbf{q,\sigma } )=\sum _{i=1}^{M}\int _{0}^{t_{i}}\mathcal{Z}_{i}^{( s)}( t)Q_{J}\mathcal{Z}_{i}^{( s)}( t)dt+\lambda _{T} \| \mathbf{T} \| +\lambda _{s} S,
\end{equation} 
where $\lambda _{s}$ are the weight of constraint penalty.

\subsection{Kinodynamic A* for initial guesses}
Directly optimizing the trajectory of a UAV payload system is challenging due to its nonlinear dynamics and time-variant shape, which impose several nonlinear constraints. Therefore, having a promising initial guess is of vital importance to prevent getting stuck in infeasible local minima and achieve faster convergence during optimization. Our planning method originates from kinodynamic hybrid A* \cite{zhou2019RobustEfficientQuadrotor}. Different from the original method regarding the quadrotor as a point, we consider the robot system's geometry shape varying in different states leveraging the differential flatness.

For hybrid A*, we use the payload acceleration $\ddot{\mathbf{x}}_L$ as the control input $u$, and represent the trajectories using 2-order time-parameterized polynomials as primitives. Each primitive is evaluated and only the collision-free ones are added to the open list. The collision checking method is similar to \ref{sec:collision_avoidance}, where we check bubbles in the grid map. 

The cost function is defined as,
\begin{equation}
    \label{equ:cost}
    C(u,T) = \int_{0}^{T} \left\Vert u \right\Vert _{2}^{2} dt + \lambda  T.
\end{equation}
$u:=\ddot{\mathbf{x}}_L$, $\lambda$ is the weight of time, and $T$ is path time. The heuristic function is the same with \cite{zhou2019RobustEfficientQuadrotor}. 

\section{NMPC with Disturbance Compensation}

In order to improve the tracking accuracy and enhance the robustness, an external force compensation strategy is employed in the Nonlinear Model Predictive Control (NMPC) for UAV payload systems to resist external forces and model uncertainties. All the perturbations can be seen as forces and torques acting on the UAV and payload. We design a force estimator cooperating with the NMPC to cancel the disturbance forces. To implicitly compensate for the torques, an Incremental Nonlinear Dynamic Inversion (INDI) method is adopted on the angular velocity controller. Our compensation mechanism can resist various disturbances, such as wind, aerodynamic drag, and even weight changes. 

\subsection{NMPC with External Force}
We extend the nominal system dynamics presented in the previous section (\ref{sec:nominalsystemdynamics}) to include an external force term acting on the UAV-payload system. The system dynamics with external force is given by the following:
\begin{align}
    \ddot{\mathbf{x}}_{Q} &=f_{c}\boldsymbol{\rho } +\frac{f\mathbf{Re}_{z}^{w} +\mathbf{f}_{Q}}{m_{Q}} - g\mathbf{e}_{z}^{w}, \label{equ:uavdynamics}\\
    \ddot{\mathbf{x}}_{L} &=-f_{c}\boldsymbol{\rho } +\frac{\mathbf{f}_{L}}{m_{L}} - g\mathbf{e}_{z}^{w}, \label{equ:payloaddynamics}\\
    f_{c} &=\frac{m_{L} l(\dot{\boldsymbol{\rho }} \cdotp \dot{\boldsymbol{\rho }})}{( m\mathcal{_{\mathnormal{Q}}} +m_{\mathnormal{L}})} +\frac{\boldsymbol{\rho } \cdot \left( m\mathcal{_{\mathnormal{Q}}}\mathbf{f}_{\mathnormal{L}}  - m_{\mathcal{L}}\left( f\mathbf{Re}_{z}^{w} +\mathbf{f}_{Q}\right)\right)}{m\mathcal{_{\mathnormal{Q}}}( m\mathcal{_{\mathnormal{Q}}} +m\mathnormal{_{\mathnormal{L}}})}, \label{equ:cabletension}\\
    \mathbf{\ddot{\boldsymbol{\rho }}} &=\boldsymbol{\rho } \times \left(\boldsymbol{\rho } \times \left(\frac{f\mathbf{Re}_{z}^{w} +\mathbf{f}\mathcal{_{\mathnormal{Q}}}}{m\mathcal{_{\mathnormal{Q}}} l} -\frac{\mathbf{f}_{L}}{m_{L} l}\right)\right) -(\dot{\boldsymbol{\rho }} \cdotp \dot{\boldsymbol{\rho }})\boldsymbol{\rho }, \label{equ:dotdotcabledirection}\\
    \dot{\boldsymbol{\rho }} &=\frac{d\boldsymbol{\rho }}{dt}, \ \dot{\mathbf{x}}_{L} =\frac{d\mathbf{x}_{\mathnormal{L}}}{dt} , \  \dot{\mathbf{x}}_{Q} =\frac{d\mathbf{x}\mathnormal{_{Q}}}{dt}, \ \dot{\mathbf{R}} =\mathbf{R}\hat{\boldsymbol{\omega }}, \label{equ:dynamics}
\end{align}
where the $f_Q$ and $f_L$ denote the external force acting on the UAV and payload, respectively. $f_c$ expressed by \eqref{equ:cabletension} is the tension on the cable and $\boldsymbol{\rho } $ is the cable direction. The UAV and payload dynamics are given by \eqref{equ:uavdynamics} and \eqref{equ:payloaddynamics}. 

The NMPC problem is written as, 
\begin{subequations}
    \begin{align}
        \underset{\mathbf{u}}{\min} \mathcal{J} &=\frac{1}{2}\sum _{k=1}^{N}\left(\left\Vert \tilde{\mathbf{x}}( k)\right\Vert _{\mathbf{Q}_{k}}^{2} +\left\Vert \tilde{\mathbf{u}}( k)\right\Vert _{\mathbf{H}_{k}}^{2}  \right) +\frac{1}{2}\left\Vert \tilde{\mathbf{x}}( N+1)\right\Vert _{\mathbf{Q}_{e}}^{2}, \\
        s.t. \  &\mathbf{x}( k+1) =\mathrm{f}(\mathbf{x}( k) ,\mathbf{u}( k)),\\
        &\mathbf{u}_{min} \leqslant \mathbf{u} \leqslant \mathbf{u}_{max}, \ \mathbf{x}( N+1) \in X_{f},
    \end{align}
\end{subequations}
where $\tilde{\mathbf{x}}( k) =\mathbf{x}( k) -\mathbf{x}_{r}( k), \tilde{\mathbf{u}}( k) =\mathbf{u}( k) -\mathbf {u}_{r}( k)$. $\mathbf{x}_{r}$ and $\mathbf{u}_{r}$ are the reference state and control input derived from trajectories. Control input $\mathbf{u}=[f_c, \boldsymbol{\omega }_c]^T$ includes command thrust $f_c$ and command angular velocity $\boldsymbol{\omega }_c$. System state is $\mathbf{x}=[\mathbf{x}_{Q}, \mathbf{x}_{L}, \dot{\mathbf{x}}_{Q},\dot{\mathbf{x}}_{L}, \boldsymbol{\rho}, \dot{\boldsymbol{\rho }}, \mathbf{R}]^T$. 
$\mathbf{u}_{min}$ and $\mathbf{u}_{max}$ are constraints of control inputs imposed by the physical and operational limitations. The function $\mathrm{f}(\mathbf{x}( k) ,\mathbf{u}( k))$ is a discrete version of system dynamics (equation \eqref{equ:uavdynamics}-\eqref{equ:dynamics}). $N$ is the prediction horizon. $\mathbf{Q}_{k}$ and $\mathbf{H}_{k}$ are the state and control input weighting matrices at $k$ step. $\mathbf{Q}_{e}$ is the terminal state weighting matrix. The $\mathbf{x}( N+1) \in X_{f}$ is the terminal constraint.
Note that, we apply exponential decay on the weight matrix $\mathbf{Q}$ and $\mathbf{H}$, written as \eqref{eq_exp_decay}, to improve the controller robustness, avoid ill condition\cite{Wang2001}. Intuitively, the decay weight prioritizes the error at the current time over future times\cite{Wang2001,Yoon1993}, reducing the impact of the estimation noise and model uncertainty. The $b_{\mathbf{x}}$ and $b_{\mathbf{u}}$ decide the decay speed. To solve the NMPC problem, we use the real-time iteration (RTI) scheme with Gauss-Newton hessian approximation. 
We implement the code with ACADO\cite{Houska2011b}.
\begin{equation}
	\label{eq_exp_decay}
	\mathbf{Q}_{k} =\exp\left( -\frac{k}{N} b_{\mathbf{x}}\right)\mathbf{Q} ,\ \mathbf{H}_{k} =\exp\left( -\frac{k}{N} b_{\mathbf{u}}\right)\mathbf{H}.
\end{equation}

\subsection{External Force Estimator}

\begin{figure}[thpb]
    \centering
    \subfigure[External force model]  {\includegraphics[width=0.46\linewidth]{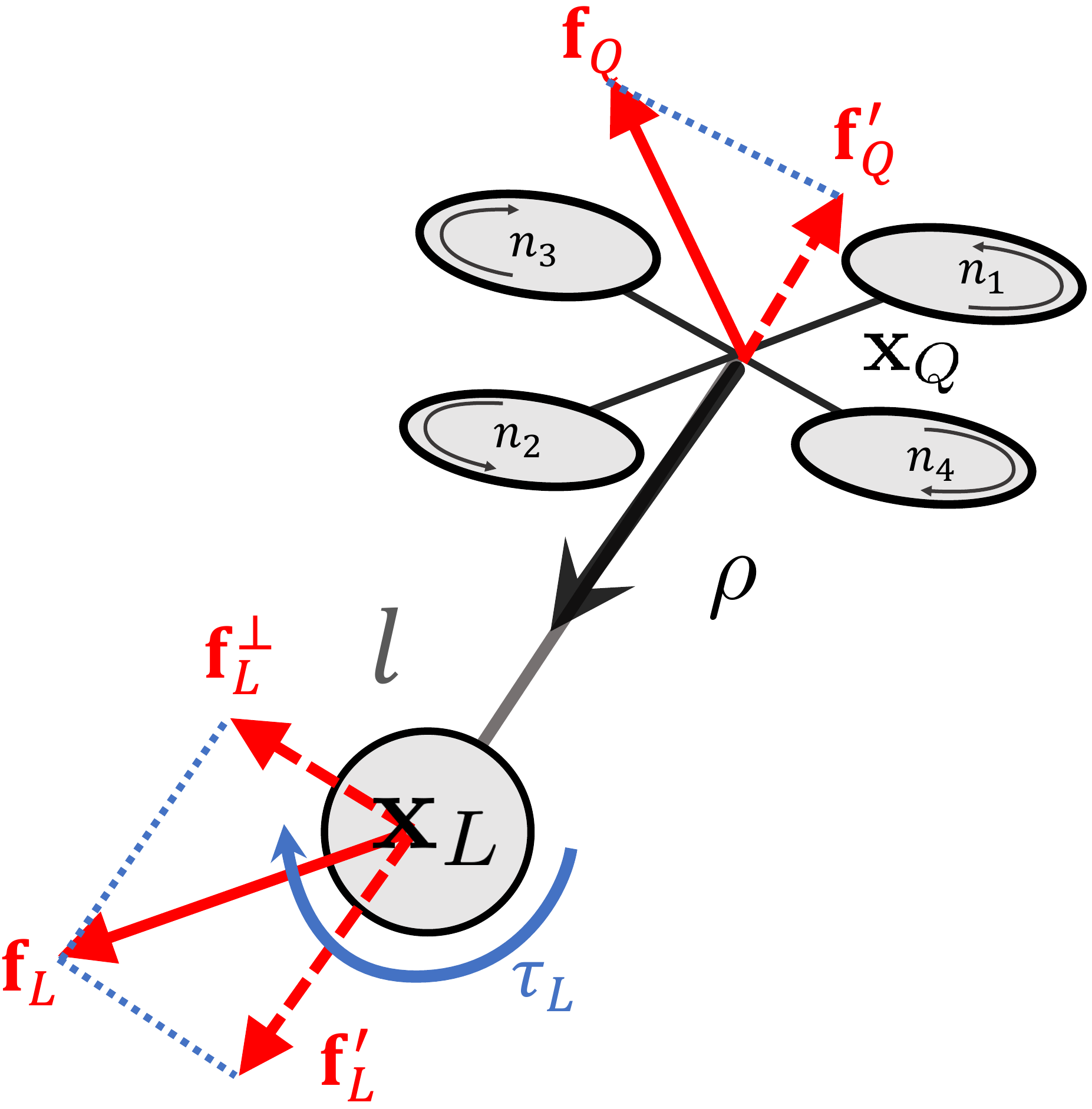} \label{fig:payload_force}}
    \subfigure[Mount point mismatch]  {\includegraphics[width=0.34\linewidth]{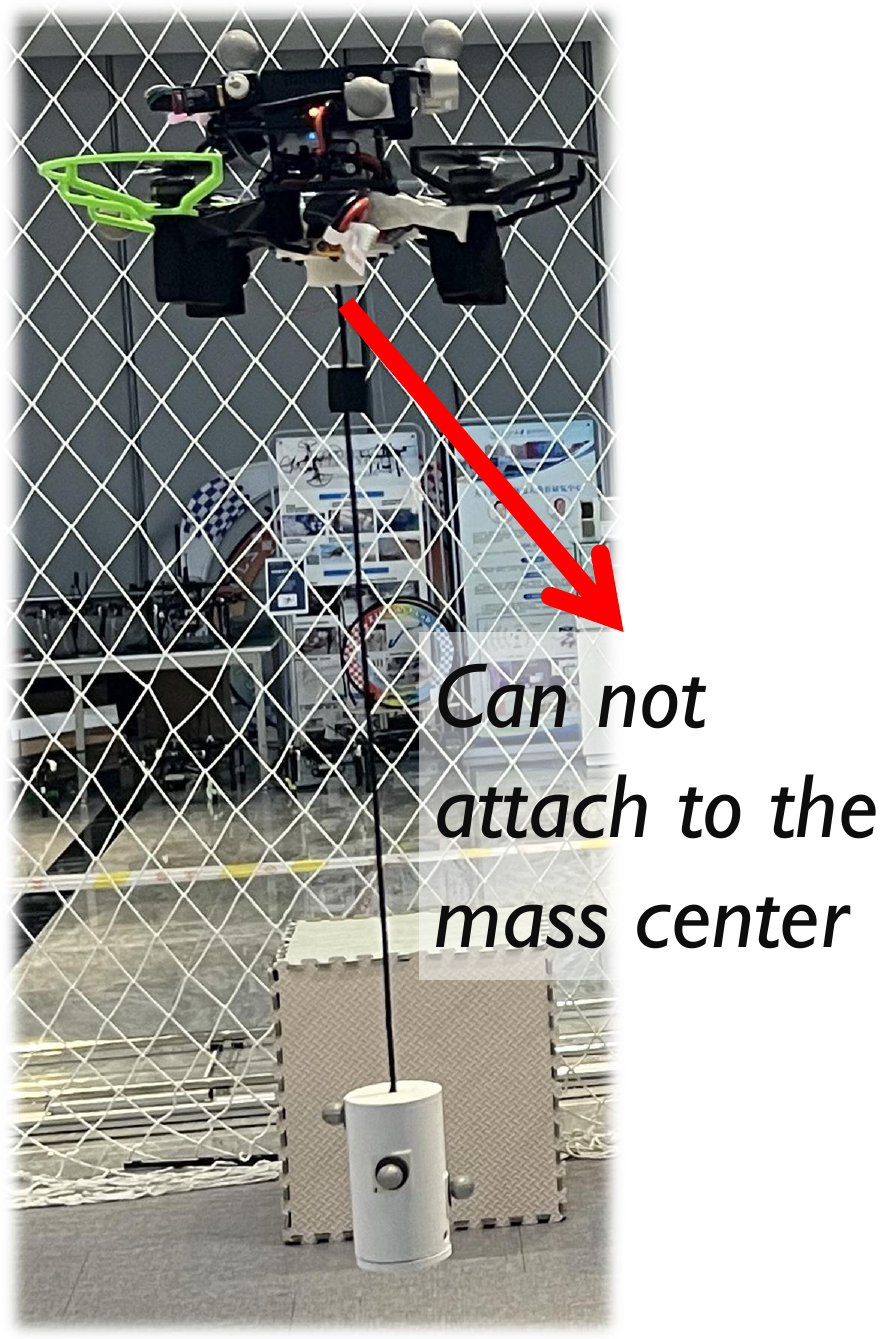} \label{fig:payload_indi}}
    \caption{(a) shows the external forces acting on the system. The component force along the direction of the cable $\mathbf{f}_{Q}^{'} $ and $\mathbf{f}_{L}^{'}$ would cancel each other. Another force $\mathbf{f}_{L}^{\bot}$ can cause the moment $\tau_L$ acting on the cable. (b) shows the mount point mismatch. In practice, It is hard to satisfy that the mount point is exactly the same as the center of mass.}
    \vspace{-0.4cm}
\end{figure}

Obtaining accurate external force is key to the proposed controller. We propose an effective online estimator to obtain the external force. 

The forces are illustrated in Figure. \ref{fig:payload_force}.  Adding equation \eqref{equ:uavdynamics} to \eqref{equ:payloaddynamics}, we can obtain 
\begin{equation}
    \mathbf{f}_{Q} +\mathbf{f}_{L}  = m_{Q}\left( g\mathbf{e}_{z}^{w} +\ddot{\mathbf{x}}_{Q}\right) +m_{L}\left( g\mathbf{e}_{z}^{w} +\ddot{\mathbf{x}}_{L}\right),
\end{equation}
where the $\ddot{\mathbf{x}}_{Q}$ and $\ddot{\mathbf{x}}_{L}$ are the acceleration of UAV and payload which can be easily measured by IMU. 

From \eqref{equ:uavdynamics}, we have
\begin{equation}
    \boldsymbol{\rho } \times \mathbf{f}_{Q} =\boldsymbol{\rho } \times \left( f\mathbf{Re}_{z}^{w} -m_{Q} g\mathbf{e}_{z}^{w} -m_{Q}\ddot{\mathbf{x}}_{Q}\right).
\end{equation}
The $f\mathbf{Re}_{z}^{w}$ is the thrust of the UAV in the world frame. The thrust $f$ can be obtained through the angular velocity of each propeller, $\mathbf{n}$ (measured by bi-directional Dshot ESC), by the propeller model of quadrotor like
\begin{equation}
    \mathbf{G}_{1} =\begin{bmatrix}
        k_{f} & k_{f} & k_{f} & k_{f}\\
        a_{y} k_{f} & -a_{y} k_{f} & -a_{y} k_{f} & a_{y} k_{f}\\
        -a_{x} k_{f} & -a_{x} k_{f} & a_{x} k_{f} & a_{x} k_{f}\\
        -k_{m} & k_{m} & -k_{m} & k_{m}
        \end{bmatrix},
\end{equation}
\begin{equation}
    \mathbf{G}_{2} =\begin{bmatrix}
        \mathbf{0_{3\times 4}}\\
        \begin{matrix}
        I_{p} & -I_{p} & I_{p} & -I_{p}
        \end{matrix}
        \end{bmatrix}, \begin{bmatrix}
            f\\
            \boldsymbol{\tau}
            \end{bmatrix} = \mathbf{G}_{1} \mathbf{n}^{\circ 2} +\mathbf{G}_{2}\dot{\mathbf{n}},
            \label{equ:propeller_model}
\end{equation}

where $a_{x} =a\cos \beta $, $a_{y} =a\sin \beta $, $a$ is the arm length of the quadrotor, $\beta $ is the angle between the quadrotor $x$ axis and the propeller, $k_{f}$ and $k_{m}$ indicate thrust coefficient and torque coefficient respectively, $I_{p}$ is the inertia of the propeller, $\boldsymbol{\tau}$ is the torque of the quadrotor, and $\circ $ represents the Hadamard power. 

It should be noted that when the payload is kept at a fixed angle to the UAV, such as the hovering state, equations \eqref{equ:uavdynamics} and \eqref{equ:payloaddynamics} are underdetermined for $\mathbf{f}_{Q}$ and $\mathbf{f}_{L}$, leading to infinite solutions. 
Physically, $\mathbf{f}_{L}$ and $\mathbf{f}_{Q}$ can be resolved into components along and perpendicular to the cable. However, the component force along the cable cancels out, leading to infinite solutions. To avoid this, a nominal combination of forces for $\mathbf{f}_{L}$ and $\mathbf{f}_{Q}$ is found. We assume that $\mathbf{f}_{L}$ is perpendicular to the cable and creates a moment $\boldsymbol{\tau}_L$ on it. Therefore, we have $\mathbf{f}_{L} \cdotp \boldsymbol{\rho} = 0$ and $\boldsymbol{\tau}_L = \boldsymbol{\rho} \times \mathbf{f}_{L} $.

We formulate the estimation problem as a nonlinear least squares problem. Residuals are defined as 
\begin{align}
    r_{1}^i &=\mathbf{f}_{Q}^i +\mathbf{f}_{L}^i -m_{Q}\left( g\mathbf{e}_{z}^{w} +\ddot{\mathbf{x}}_{Q}^i\right) +m_{L}\left( g\mathbf{e}_{z}^{w} +\ddot{\mathbf{x}}_{L}^i\right), \\
    r_{2}^i &=\boldsymbol{\rho }^i \times \mathbf{f}_{Q}^i -\boldsymbol{\rho }^i \times \left( f^i\mathbf{R}^i \mathbf{e}_{z}^{w} -m_{Q} g\mathbf{e}_{z}^{w} -m_{Q}\ddot{\mathbf{x}}_{Q}^i\right), \\
    r_{3}^i &=\mathbf{f}_{L}^i \cdotp \boldsymbol{\rho}^i,
\end{align}
where the $i$ denotes each measurement. For robustness, we utilize a sliding window to estimate the forces in our implementation. Assuming that external forces vary insignificantly within the sliding window, we add a variance regularization term. The objective function is expressed as
\begin{equation}
    \underset{\mathbf{f}_{Q} ,\mathbf{f}_{L}}{\min}\sum _{i=1}^{N}\left( L_{\delta }\left( r^{i}\right) +k_{r} L_{\delta }\left(\mathbf{f_{Q}^{i}} -\overline{\mathbf{f_{Q}}}\right) +k_{r} L_{\delta }\left(\mathbf{f_{L}^{i}} -\overline{\mathbf{f_{L}}}\right)\right),
\end{equation}
where $L_{\delta }$ is Huber loss, $k_r$ is the regularization parameter, $\overline{\mathbf{f_{Q}}}$ and $\overline{\mathbf{f_{L}}}$ are the mean of the forces in the sliding window. We solve this problem by L-BFGS\cite{liu1989Limited}.

To incorporate with NMPC, we assume the external forces $\mathbf{f_{Q}}$ and $\mathbf{f_{L}}$ in system dynamics (equation \eqref{equ:uavdynamics}-\eqref{equ:dynamics}) keep constant during the prediction horizon. The external forces are defined as online parameters of the system dynamics model. Our estimator updates the $\mathbf{f_{Q}}$ and $\mathbf{f_{L}}$ at a high frequency (100Hz), allowing the controller to continuously adapt to the changing system dynamics and external conditions.

\subsection{INDI Inner Loop Controller}

In the previous sections, we always assume the payload is attached to the UAV mass center. However, in practice, this assumption is hard to satisfy, as shown in Figure. \ref{fig:payload_indi}. This leads to an additional torque on the UAV. Measuring the distance between CoM and attached point and modeling the associated torque is non-trivial. Inspired by \cite{tal2021AccurateTrackingAggressive,sun2022ComparativeStudyNonlinear}, we adopt an Incremental Nonlinear Dynamic Inversion-based inner loop controller to eliminate this torque implicitly.

After getting the angular velocity commands $\boldsymbol{\omega}_{r}$ from previous NMPC controller, The desired angular acceleration $\boldsymbol{\dot{\omega}}_{c} $ is calculated as 
\begin{equation}
    \boldsymbol{\dot{\omega}}_{c} =K_{\omega}( \boldsymbol{\omega}_{r} -\boldsymbol{\omega}_{f}) +\boldsymbol{\dot{\omega}}_{r},
\end{equation}
where $\boldsymbol{\omega}_{f}$ is the filtered feedback angular velocity from quadrotor IMU, $\boldsymbol{\dot{\omega}}_{r}$ denotes reference angular acceleration. The INDI control law\cite{tal2021AccurateTrackingAggressive} is then applied to calculate the control moment $\boldsymbol{\tau}_{c}$. 
\begin{equation}
\boldsymbol{\tau}_{c} =\boldsymbol{\tau}_{f} +\mathbf{J}\left( \boldsymbol{\dot{\omega}}_{c} - \boldsymbol{\dot{\omega}}_{f}\right), \label{equ:indi}
\end{equation}
where the $\boldsymbol{\tau}_{f}$ is the filtered feedback angular moment obtained through the angular velocity of each propeller based on the propeller model (equation \eqref{equ:propeller_model}) and $\boldsymbol{\dot{\omega}}_{f}$ is the feedback angular acceleration. The term $\boldsymbol{\tau}_{f} - \mathbf{J}\boldsymbol{\dot{\omega}}_{f} $ in \eqref{equ:indi} can implicitly compensate the unknown torque on the UAV \cite{tal2021AccurateTrackingAggressive}. Then we solve the following equation to obtain each desired rotor angular velocity $n_c$. Here, $f_{tc} $ is desired thrust.
\begin{equation}
    \begin{bmatrix}
        f_{tc}\\
        \boldsymbol{\tau}_{c}
        \end{bmatrix} =\mathbf{G}_{1}\mathbf{n}_{c}^{\circ 2} +\Delta t_m^{-1}\mathbf{G}_{2}(\mathbf{n}_{c} -\mathbf{n}_f),
\end{equation}
where $\Delta t_m$ is the motor time constant and $\mathbf{n}_f$ is the current rotor angular velocity. In the implement, $\boldsymbol{\omega}_{f}$ and $\mathbf{n}_f$ are filtered with the second-order Butterworth filter with a same cut-off frequency (10Hz) to ensure that they have a similar phrase and delay. This is important for sensors' synchronization. The $\boldsymbol{\dot{\omega}}_{f}$ is obtained from noise-robust numerical derivative to avoid IMU's noise being amplified.

\section{Experiment and Evaluation}

\begin{figure}[thbp]
    \centering
    \includegraphics[width=0.95\linewidth]{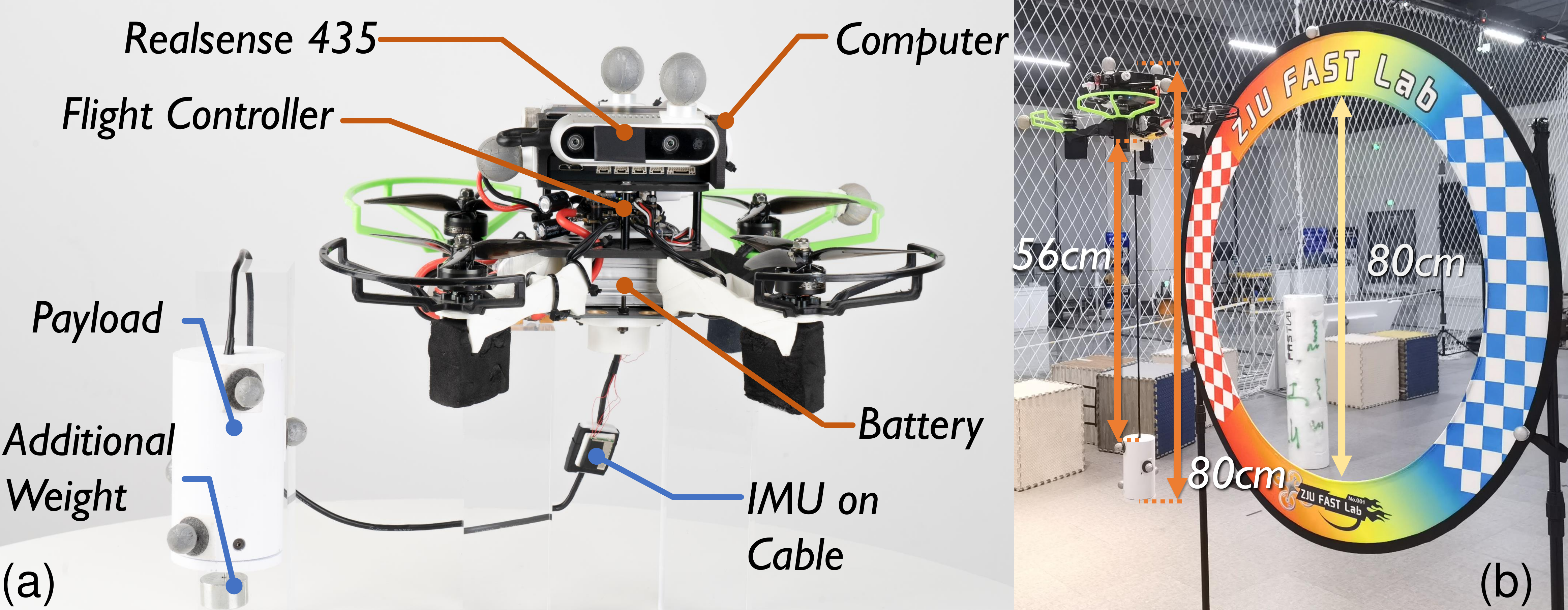}
    \caption{(a) Quadrotor with suspended payload platform. Additional weight is hooked on the payload. (b) Gate in aggressive flight experiment.}
    \label{fig:exp_setup}
    \vspace{-0.35cm}
\end{figure}

We evaluate the performance of the proposed planning and control methods through adequate simulation and real-world experiments. In this section, we aim to answer three questions. i) Can the planning module generate dynamic feasible, and collision-free trajectories in clutter environments while meeting the real-time requirement? (See \ref{sec:exp_planning}.) ii) Can the controller resist the disturbance and the model parameter inaccurate, achieving "load-and-fly"? (See \ref{sec:exp_controller}.) iii) What is the performance of the whole framework in the real world? (See \ref{sec:exp_aggressive_flight}, \ref{sec:exp_unknown_env} and our video.)

We set up our experiments with a customized quadrotor platform, shown in \autoref{fig:exp_setup} (a). The platform consists of an Ardupilot flight controller (Kakute H7 stack), an onboard computer (DJI Manifold 2C), and an Intel RealSense D435 camera. The payload has a mass of 285g, with the capability to add additional weight. The quadrotor's propeller angular velocities are obtained through bi-directional dshot ESC. Accurate positioning information for quadrotor and suspended payload is provided by a motion capture system (MoCap). We fuse the data from IMU on the cable and MoCap to estimate the direction of the cable and payload's position, velocity by extend kalman filter. The IMU data is used for prediction, while the MoCap provides high-precision correction. All the modules, including planning, controller, and perception, run on the onboard computer.

\subsection{Planning Performance}
\label{sec:exp_planning}
\begin{figure}[t]
    \centering
    \includegraphics[width=0.95\linewidth]{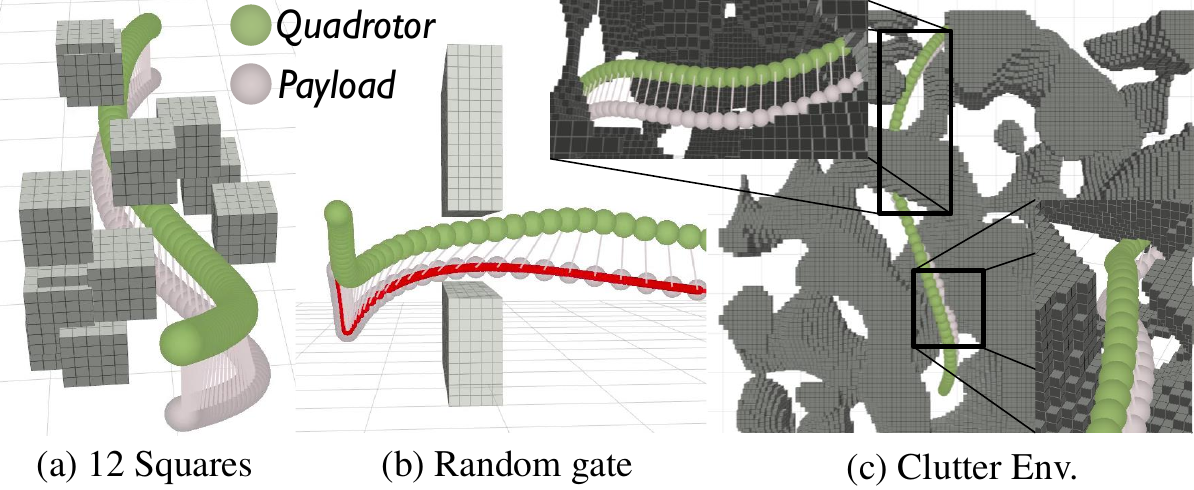}
    \caption{Planning experiments in three different scenarios. }
    \label{fig:planning_sim}
    \vspace{-0.2cm}
\end{figure}

To answer the first question, we evaluate the planning module in three simulation scenarios, like \autoref{fig:planning_sim}, and compare its average computational time and successful rate against other planning methods\cite{foehn2017FastTrajectoryOptimization,son2020Realtime}. Due to no source code available, we implement other methods with CasADi\cite{Andersson2019} by ourselves. The first scene is a simple 3D environment containing 12 random squares. The second scene is passing a random gap. The third scene is a complex and cluttered environment. We created 10 simulation environments for each scenario to measure the success rate. It should be noted that Son's method \cite{son2020Realtime} does not support non-convex environments, so we only tested it in the first two scenarios.

\autoref{tab:planning} illustrates our method achieved the highest success rate of 97.5\% while offering a significant improvement in computation time compared to other methods, as our method performed hundreds of times faster. 

\begin{table}
    \centering
    \caption{Planning Performance Comparison}
    \label{tab:planning}
    \begin{tabular}{ccccccc} 
    \toprule
          & \multicolumn{2}{c}{12 Squares} & \multicolumn{2}{c}{Random 
          Gap} & \multicolumn{2}{c}{Clutter Env.}  \\ 
    \cline{2-7}
          & Time[s] & Succ.              & Time & Succ.               & Time & Succ.                           \\ 
    \midrule
    Ours  &    \textbf{0.021}  &      \textbf{100\% }                &     \textbf{0.017}&      \textbf{100\% }                  &    \textbf{0.063}   &          \textbf{90\% }                           \\
    Foehn\cite{foehn2017FastTrajectoryOptimization} &  17.1    &   90\%  &   15.8   &   80\%              &  54.8    &        40\%                              \\
    Son\cite{son2020Realtime}   &   0.437   &  80\%                       &    2.2  &       50\%                  &    - &    -                                 \\
    \bottomrule
    \end{tabular}
    \vspace{-0.45cm}
    \end{table}

\vspace{-0.15cm}
\subsection{NMPC Ablation Study}
\label{sec:exp_controller}

\begin{figure}[t]
    \centering
    \includegraphics[width=0.95\linewidth]{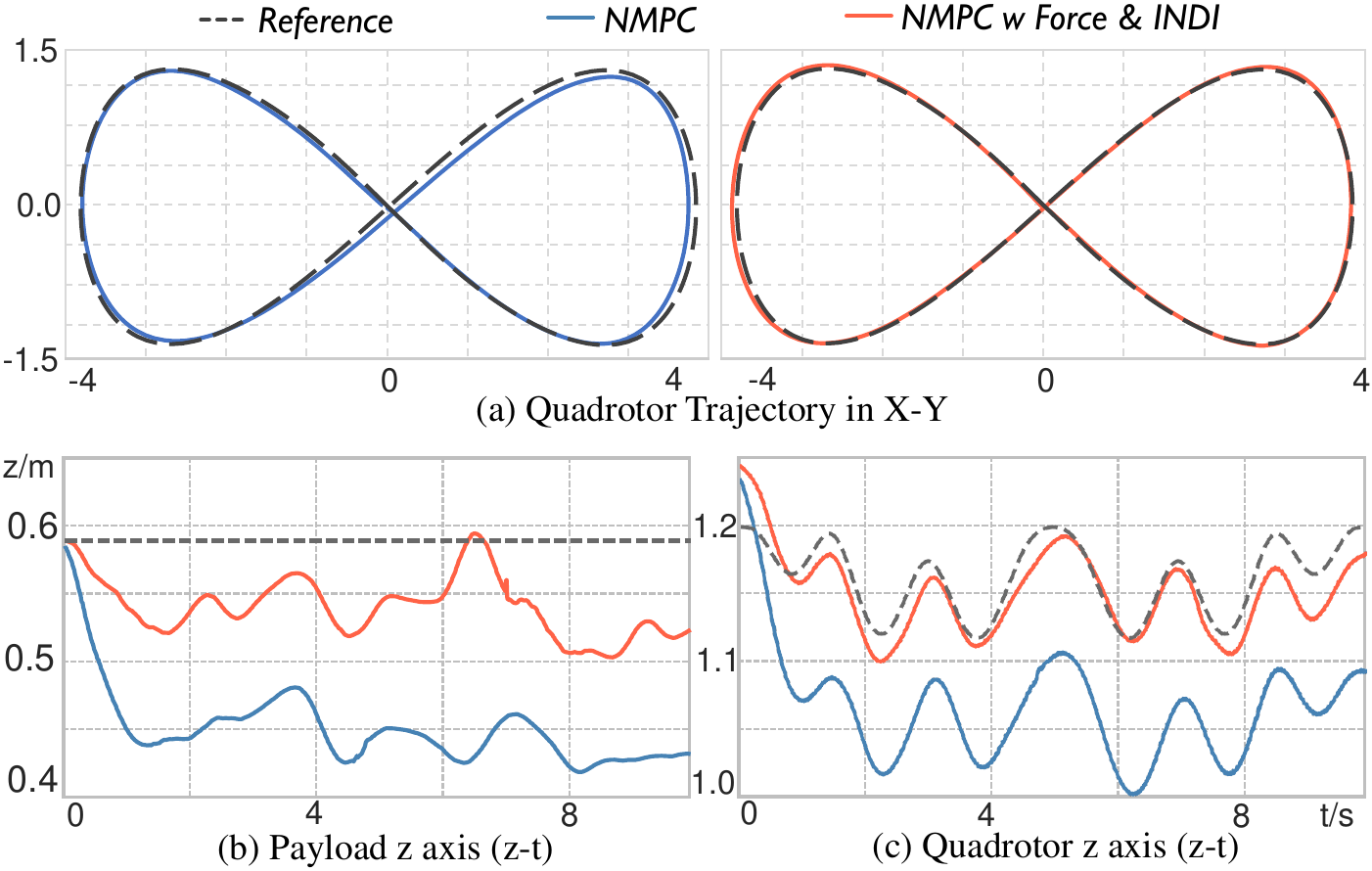}
    \caption{Position Tracking Comparison with 200g unknown weight. (a) shows quadrotor tracking error in X-Y plane. (b),(c) show the payload and quadrotor tracking error in Z axis. NMPC with force compensation and INDI shows better tracking performance than only NMPC. }
    \label{fig:MPC_traj}
    \vspace{-0.2cm}
\end{figure}

To answer the second question, we evaluate the proposed controller on the same eight trajectory like \autoref{fig:MPC_traj}. 
We evaluate the controller's basic performance with different speed profiles and assess its performance against parameter uncertainty by introducing two unknown weights (50g and 200g). To validate its performance in unfamiliar conditions, we create two different wind environments using a fan, with wind speeds of 4.5 m/s and 9 m/s. The performance is measured as RMSE/MAX in \autoref{tab:NMPCAblation}.

\begin{figure}[t]
    \centering
    \includegraphics[width=0.95\linewidth]{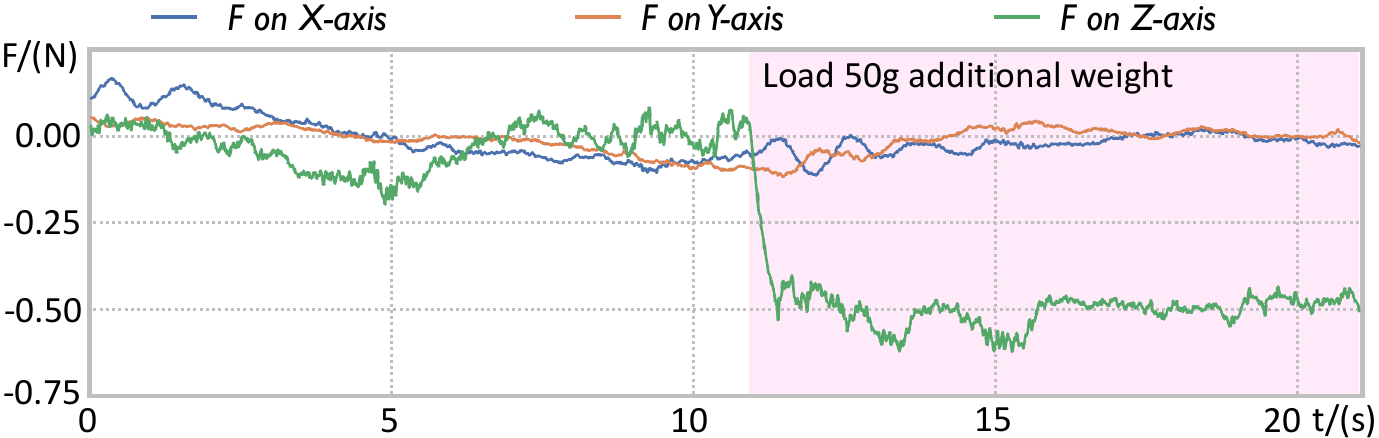}
    \caption{ Force estimation results ($\mathbf{f}_{Q}$) when we load a 50g additional weight.}
    \label{fig:weightchange}
    \vspace{-0.5cm}
\end{figure}

\begin{table*}[t]
    \centering
    \caption{NMPC Ablation Study in Real Environment}
    \label{tab:NMPCAblation}
    \begin{tabular}{ccc|cccc|cccc}
    \toprule
    \multicolumn{3}{c|}{Trajectory \& Environment} &
      \multicolumn{4}{c|}{Quadrotor Position Tracking Error (RMSE/MAX) {[}cm{]}} &
      \multicolumn{4}{c}{Payload Position Tracking Error (RMSE/MAX) {[}cm{]}} \\ \midrule
      \begin{tabular}[c]{@{}c@{}}$v_{max}$\\ {[}$m/s${]}\end{tabular}  &
      \begin{tabular}[c]{@{}c@{}}$a_{max}$\\ {[}$m/s^2${]}\end{tabular} &
      Condition &
      NMPC &
      w INDI &
      w Force &
      \begin{tabular}[c]{c} w Force\\ \& INDI\end{tabular} &
      NMPC &
      w INDI &
      w Force &
      \begin{tabular}[c]{c} w Force\\ \& INDI\end{tabular}\\ 
      \midrule
      1.5                                                         & 2.1                                                         & \multirow{4}{*}{None} & 5.2/9.1   & 4.9/7.9   & 3.4/6.5  & \textbf{2.4/5.0}       & 8.4/13.7  & 9.3/14.1  & 6.8/10.9  & \textbf{4.4/8.8}    \\
      2.5                                                         & 3.7                                                         &                       & 7.1/11.5  & 7.1/11.7  & 4.4/6.9  & \textbf{3.9/6.5}       & 10.0/21.7 & 9.9/21.1 & 8.7/13.1  & \textbf{5.6/10.9}   \\
      4.0                                                         & 5.7                                                         &                       & 7.7/12.8  & 7.6/13.3  & 5.1/9.9  & \textbf{4.9/9.7}       & 10.2/22.1 & 11.9/25.5 & 9.4/19.7  & \textbf{7.7/12.1}   \\
      5.9                                                         & 9.0                                                         &                       & 9.0/14.5  & 8.7/15.3  & 7.9/17.3 & \textbf{7.1/15.0}      & 14.6/27.2 & 14.6/26.7 & 12.1/27.1 & \textbf{11.5/22.0}  \\ 
      \midrule
      \multirow{4}{*}{4.0}                                        & \multirow{4}{*}{5.7}                                        & Weight +50g           & 8.0/12.6  & 6.8/11.4  & 5.8/10.2 & \textbf{5.4/10.5}      & 13.3/24.3 & 12.2/21.1   & 10.2/19.9 & \textbf{9.5/15.3}   \\
                                                                  &                                                             & Weight +200g          & 12.4/15.5  & 11.8/15.2 & 6.4/11.5 & \textbf{5.5/11.0}      & 19.5/27.0 & 18.0/25.0    & \textbf{10.8/17.1} & 10.9/17.1  \\
                                                                  &                                                             & Wind 4.5m/s           & 8.0/14.3  & 7.3/11.8  & 6.0/11.3 & \textbf{5.4/11.1}      & 11.1/23.3 & 10.5/18.1   & 9.7/19.9 & \textbf{9.0/15.5}   \\
                                                                  &                                                             & Wind 9m/s             & 8.2/13.6  & 7.2/11.6  & 6.8/11.7 & \textbf{5.5/10.0}      & 11.9/22.0 & 10.5/18.2   & 9.8/21.3 & \textbf{9.2/20.3}   \\
      \bottomrule
    \end{tabular}
    \vspace{-0.45cm}
\end{table*}

The results illustrate that NMPC with force compensation and INDI has the best performance. Overall, the NMPC with force compensation and INDI can reduce about 35\% tracking error. As \autoref{fig:MPC_traj}, when we hung an additional 200g weight on the payload, the NMPC with force compensation and INDI can still keep the similar performance with no additional weight while without the force compensation and INDI, the tracking error increases significantly. 

Furthermore,  We add a 50g additional weight when the quadrotor is flying to verify the force estimator's performance. As shown in \autoref{fig:weightchange}, when we load the weight, the z-axis of $\mathbf{f}_{Q}$ changes from 0N to about -0.49N. This enables load-and-fly operation where the payload can be attached to the quadrotor without prior knowledge of its weight. 

\subsection{Aggressive Flight Experiment}
\label{sec:exp_aggressive_flight}
We evaluate the whole framework in an aggressive flight experiment, where the system has to pass through a gate, shown as \autoref{fig:exp_setup} (b). In this experiment, we aim to demonstrate the ability of our planning method to generate dynamically feasible trajectories in real time, even in challenging scenarios. The controller can resist unknown weight added to the payload, even with maneuvers.  

In the experiment, The UAV payload system, even with an unknown weight (50g), passed through the gate multiple times, with the gate's position varied for each repetition. The maximum velocity and acceleration of the payload were measured at $5.2m/s$ and $7.3m/s^2$, respectively. The system exhibited the capability to navigate through the gate in real time while performing the necessary maneuvers.

\subsection{Autonomous Navigation in Unknown Environments}
\label{sec:exp_unknown_env}

\begin{figure}[t]
    \centering
    \includegraphics[width=0.9\linewidth]{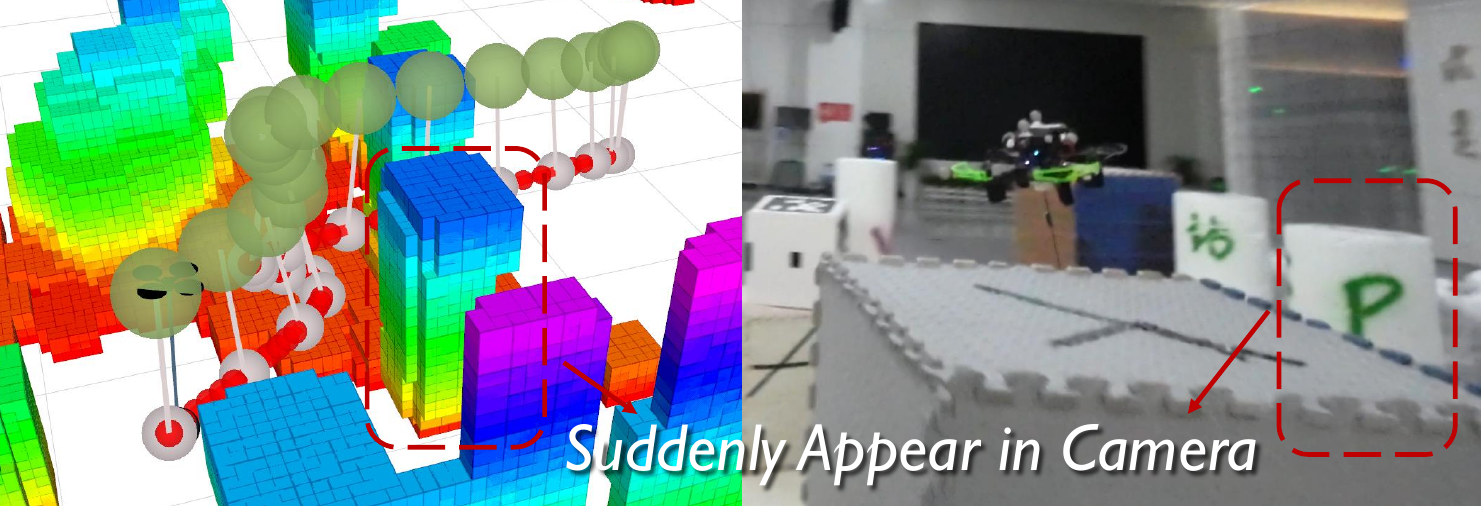}
    \caption{The UAV-payload system performs autonomous navigation in complex unknown environments. It can fast modify the trajectory to handle the sudden appearance of new obstacles due to the limited perception range. }
    \label{fig:unknown_env}
    \vspace{-0.55cm}
\end{figure}
To further validate the robustness of our method, we conduct autonomous navigation experiments in complex unknown environments. The UAV payload system was equipped with an onboard camera for environment perception. The motion planning module was challenged by the complex surroundings and limited perception range, requiring rapid and continual trajectory re-planning in response to sudden obstacles to ensure safety. Three different experiment cases were set up, and some shots are presented in \autoref{fig:unknown_env}. The full experiments can be seen in the accompanying video.

\section{Conclusion}
This paper proposes a practical solution for autonomous UAV payload system navigation. We proposed an efficient trajectory generation formulation meeting safe and real-time requirements and an NMPC controller with a hierarchical disturbance compensation strategy to resist disturbances. The sufficient experiments demonstrate the effectiveness and superiority of our method. Currently, we assume the external forces remain constant for a short time, which may result in a mismatch. Our future work will focus on incorporating a small neural network to predict external disturbances.

\vspace{-0.2cm}









\bibliography{references}

\end{document}